# Title: Landscape of Machine Implemented Ethics

*Vivek Nallur (vivek.nallur@ucd.ie)*

*University College Dublin, Ireland*

**Abstract**: This paper surveys the state-of-the-art in machine ethics, that is, considerations of how to implement ethical behaviour in robots, unmanned autonomous vehicles, or software systems. The emphasis is on covering the breadth of ethical theories being considered by implementors, as well as the implementation techniques being used. There is no consensus on which ethical theory is best suited for any particular domain, nor is there any agreement on which technique is best placed to implement a particular theory. Another unresolved problem in these implementations of ethical theories is how to objectively validate the implementations. The paper discusses the dilemmas being used as validating 'whetstones' and whether any alternative validation mechanism exists. Finally, it speculates that an intermediate step of creating domain-specific ethics might be a possible stepping stone towards creating machines that exhibit ethical behaviour.

***Contact:*** School of Computer Science, University College Dublin, Dublin, Republic of Ireland. Eircode: D04 V1W8. Tel: +353 (1) 7162475. Email: vivek.nallur@ucd.ie

## Motivation

Computers are increasingly a part of the socio-technical systems around us. Domains such as smart-grids, cloud computing, healthcare, and transport are but some examples where computers are deeply embedded. The speed and complexity of decision-making in these domains have meant that humans are ceding more and more autonomy to these computers (Nallur & Clarke 2018). Autonomy, in machines, can be defined as the effective decision-making power over goals, that influences some action in the real-world. For instance, smart traffic lights can autonomically change their timings, depending on the flow and density of traffic on the roads. The introduction of progressive levels of autonomy into software-enabled devices that interact actively with human beings implies that human society will be impacted by decisions made by such machines. Transport systems that decide on prices of tickets based on demand (Masoum et al. 2011), smart buildings that decide how much energy to be used at a particular time (Yoon et al. 2014), smart cameras that decide which persons to track (Lewis et al. 2014), cars that can change their routing priorities (Song et al. 2015), hospital machines that recommend a particular course of treatment (Lynn 2019), are all examples of machines being given autonomy, while decisively impacting human life. Autonomous machines, therefore, need to be imbued with a sense of ethics that reflect the social milieu they operate in and make decisions that are ethically acceptable to society.

The notion of a general-purpose intelligence has been the quest of computer scientists ever since the dawn of computing. From Turing's original essay on intelligence to recent developments in machine-learning where computers outperformed humans on subtle games (AlphaGo defeated two of the greatest Go players in the world in 2016/17) and in lateral thinking (DeepMind triumphed in Jeopardy), computer science has come far. As machine-learning and self-adaptation techniques increase in sophistication, many more domains will be introduced to autonomic systems. However, the increasing pervasiveness of autonomic systems also brings uncertainty with it. System designers find that they are unable to foresee all the situations that their systems will encounter, and that interaction with other autonomic systems (humans, animals or machines), lead to entirely unpredictable results. In such scenarios, it is essential to provide basic guarantees about the kinds of behaviour exhibited by autonomic systems. Human society will likely be more willing to trust an autonomic agent if it is known to possess a set of moral principles that guide and constrain its behaviour (Bonnefon et al. 2016). There have been some attempts to insert ethical rules of behaviour into robots or other autonomous agents. Most notably, these have either been implementations of Asimov's Three Laws of Robotics (Asimov 1950) or mechanisms to deal with ethical dilemmas, such as the Trolley problem or Prisoner's Dilemma (Bjorgen et al. 2018).

## Landscape of Implementations

### *Types of Ethics*

Autonomous systems are increasingly being used in domains that have life-altering consequences. For example, the use of autonomous robots to target and kill individuals (Krishnan 2009) greatly impacts human life. A less dramatic, but still consequential example is the use of autonomic computing in healthcare, specifically in the care of elderly patients (Sharkey & Sharkey 2012). This impact on human life causes us to (currently) insist on a human-in-the-loop for ultimate decision-making. However, as multiple domains interact, and at multiple time-scales, adding humans to the loop may not be scalable. For example, emergency vehicles may use autonomic decision making to select the best hospital, and the best route to reach that hospital. In the case of civil strife or large accidents, adding humans to the decision-making loop could slow down the rate of rescue and treatment. As system designers develop such autonomic systems, it may not be evident that an action could result in a morally dubious consequence. For instance, should an autonomic system prioritise a wounded adult with a higher chance of recovery, or should it attempt to ensure that children, regardless of their state, are attended to first? Depending on the system designers' notion of ethics, there are several choices of ethical frameworks to adopt. The simplest (for the layperson to understand), and arguably the most famous, ethical framework is Asimov's *Laws of Robotics* (Asimov 1950). Isaac Asimov introduced the *Three Laws of Robotics* in his science-fiction stories, as rules that every robot was programmed with. These rules (elucidated in a later section) were meant to ensure that robots, no matter how sophisticated or powerful they became, would always be subservient to humans. No robot could take any action that would, as a consequence, result in violation of these rules. Apart from the consequentialist ethics espoused by Asimov, deontological ethical approaches such as *Prima Facie Duties* (Ross, 1987) have also been explored. Due to the emphasis on validation in computer science, and robotics, there has been more discussion about how implementations were to be evaluated, rather than on which ethical theory is a better candidate for implementation. Many implementations use ethical dilemmas as a validation proxy, *i.e.*, if the implementation can resolve a dilemma in a particular manner, then it is deemed to be a successful implementation of ethics in the robot/software agent.

This article first considers the software implementation techniques used by some researchers in various domains and then lists the ethical dilemmas that have been used for validating these implementations.

## *Techniques Used for Implementing Ethics*

The primary focus among computer scientists towards implementing ethics has been to create techniques that will regulate an agent's behaviour towards other agents, according to some norm. A norm is a behavioural constraint on an agent that regulates and structures the social order within a multi-agent system. Agreement on a norm helps to promote cooperation and coordination between heterogeneous agents in open systems. If the set of norms in an agent-system can be shown to be both, individually and collectively ethical, then the behaviour of the agents, though autonomous, can also be expected to be ethical as long as they follow one/some of the norms. Which norms to follow in case of a difference of opinion between two or more agents, is an open question. Unfortunately, the difficulty of getting multiple autonomous agents to agree on even a single norm (Kittock 1993) has been shown to be hard. One of the first attempts at getting an agent to be explicitly ethical was attempted by implementing ethical rules in the medical domain (Anderson et al. 2006). The authors implemented a version of ethical theory, that is tailored to the medical domain, specifically Beauchamp's and Childress' Principles of Biomedical Ethics (Beauchamp & Childress 1991). Instead of trying to evaluate whether an agent could be programmed to be ethical, they programmed the software agent to derive generalized rules from training cases. The goal was to derive rules that would be acceptable to trained biomedical ethicists. When faced with an ethical dilemma, instead of performing any actions, the software agent provided advice, along with reasoning based on the rules it followed. For example, if a patient refuses to take an antibiotic that could potentially cure their infection, and thus save their life, the ethically accepted course of action is to try and change the patient's mind. In this case, the patient's initial decision can be considered as less than fully autonomous, since they might not be able to foresee the consequences of their action. Therefore, the software would de-prioritize patient autonomy and advise the human nurse to try again to change the patient's mind. The human nurse was free to accept or ignore this advice. The system was designed as a proof-of-concept and never tested *in-situ* with human users.

A rule-based governor is a constraint module that either allows or disallows actions to be taken by a particular system, based on whether the action would break pre-existing rules. These are typically used to ensure that autonomic systems do not break hard-constraints on their behaviour, since any plan-of-action made by the agent is first submitted to the governor for validation. If the governor approves the plan, then the agent is able to go ahead and implement the actions that constitute the plan. *Rule-Based* Ethical Governors were first implemented using formal logic (Bringsjord et al. 2006). However, the authors found that apart from problems in speed and efficiency (which can be remedied as computers get faster), handling even simple contradictions in situations and rules leads to guarantees about behaviour being rendered challenging to achieve. Another significant problem with rule-based systems is that the designer must *a priori* decide which rules must be implemented, and which can be left out. Again, not only does this decision have efficiency concerns, but also more fundamental concerns about whether the resultant agent is ethically complete or not. By ethically complete, we mean that the agent can ethically deal with *all* situations, rather than some subset that the designer anticipated. An autonomous agent must be able to deal with situations that its designer has not anticipated. Hence, if a system works on the basis of *a priori* rules, it would find it difficult to cope with novel situations. Asimov's Laws of Robotics while residing firmly in the popular imagination about ethical robots co-existing with humans, have been generally accepted to be unsuitable for actual implementation (Arkin 2008; Anderson 2011).

*Constraint-Satisfaction* techniques are reasoning techniques that attempt to check whether a proposed action would satisfy a specific set of constraints. These constraints could pertain to the state of the agent, the state of the world, or the behaviour of the agent. Unlike rules, which typically specify what actions can be taken in what context, constraints specify what *must-not-happen*. The advantage of using constraints is that if a certain constraint can be satisfied, then there is a mathematical certainty that a particular state will never be reached. These are useful when the cost of breaking the constraint in extremely high. Such constraints have been used to implement ethical behaviour in robots, particularly those that have the capability to exhibit lethal force, so that we can be guaranteed that robots will always obey the Laws Of War (Arkin 2008; Mackworth 2011). The Laws of War are derived from Just War Theory (Lazar 2017), which attempts to create a moral framework for the *why, when* and *how* war should be waged. Particular importance is attached to the principles regarding resort to war *(jus ad bellum)* and conduct during war (*jus in bello).* At least for the near future, human beings will retain control over the decision of whether to go to war or not. Hence, from an autonomous systems perspective, the most crucial aspect of the Laws of War is *conduct during war.* We would like to ensure that any autonomous weapons system deployed on the battlefield obeys these principles. The prime advantage of deploying autonomous weapon systems is that it keeps the human troops (of the deploying side) out of harm's way, while at the same time being autonomously able to make extremely quick decisions about which targets to attack, and how to attack. Depending on the constraints built into the autonomous weapon system, it could behave satisfactorily in many conditions without requiring explicit direction from the human. Using constraint-satisfaction techniques also has the advantage of being able to explain which constraint(s) prevented it from achieving a certain goal or taking a certain action. However, certain principles (e.g. discrimination among military and non-military objectives, and proportionality of harm caused in relation to achieved objective) are fundamentally difficult to assess and therefore derive constraints for. No known ethical governor would be able to perform the inference required to assess whether a response is proportional or not.

*Formal Verification* refers to using mathematical techniques to establish whether specific properties about the system can be mathematically verified. The core idea being that once a property was mathematically verified, any deviation would be an engineering mistake and therefore amenable to correction. In particular, a technique called *Model Checking* was used to provide guarantees about whether an autonomous aircraft would create and execute plans that involved ethical decision making (Dennis et al. 2016). Model Checking involves first creating a mathematical model of the system, and its environment, and then checking whether any change, due to an action, could be verified to meet certain criteria. This technique allows a system designer to work at a higher level of plans, instead of individual actions. In Dennis et al. 2016, model checking was used to verify that any plan selected by the planner did not violate any (or if it had to, violate the fewest) ethical concerns. This has the advantage of formally proving that any course of action selected would be the one that caused the least ethical concerns. Attractive in principle, this mechanism, however, assumes that there is only one context in which the agent operates (which has been foreseen) and there is only one possible plan that could apply. In unconstrained, dynamic environments, this is clearly unsuitable as indeed most ethical dilemmas suggest that more than one plan could be applied.

As opposed to pre-planning the correct course of action, *Reinforcement Learning* is a technique that learns through feedback from the environment to adjust its future behaviour. Given enough feedback from the environment (in the form of rewards), the agent can start from scratch and create a policy that adjusts the agent's behaviour in accordance with the real world. This method of learning-by-doing has been suggested as a mechanism (Abel et al. 2016) for an agent to learn the correct ethical response in a given situation. The agent solves a partially observable Markov Decision Process[1] (POMDP) to find a policy that maximizes the expected reward, given the initial state of the world, the actions the agent can take, and the partially observable environment it is in. While learning a behaviour is interesting in that it does not commit the system-designer to a definite ethical stance, the designer is still limited by having to design ethical utility functions that can be expressed in the observation function of the agent. That is, since the learned behaviour is derived from what the agent can observe, the designer has to ensure that an ethical behaviour can also be, at least potentially, derived from the agent's observations. This is complicated by the fact that even in domains where the agent can partially foresee the future, it has been shown that the correct behaviour could be computationally impossible to achieve (Mundhenk et al. 2000). This greatly limits the complexity of situations that the agent can conceivably handle.

[1] A Markov Decision Process is a mathematical framework for modelling partially random processes. It allows us to model the possible future states of an agent, given its current state and the probabilities of possible successor states.

An approach that incorporates multiple ethical theories, instead of trying to pick one particular solution, has been attempted in a project called HERA (short for *hybrid ethical reasoning agents*) (Lindner et al. 2017). In this approach, ethical principles are modelled as logical formulae. Depending on whether certain formulae can be shown to be true/false, actions can be permitted or not. To achieve this, it models actions and their consequences as directed acyclic graphs, which allows the system to reason about which actions could lead to what consequences. This approach has the advantage of being much more ethically flexible than other attempts, since different philosophical approaches can be modelled, at the same time. However, an important concern is the need for a human to engineer or pre-create causal models for the agent, *i.e.*, an autonomous agent cannot modify its causal graphs in the light of new information, new contexts, or new environments.

While it is not possible to give a detailed account of each implementation technique, it would be instructive to go deeper into one implementation. Let us consider the most straightforward consequential framework: Asimov's Laws of Robotics. The Three Laws of Robotics (Asimov 1950) can be elucidated as follows:

1. A robot may not injure a human being or, through inaction, allow a human being to come to harm
2. A robot must obey the orders given it by human beings except where such orders would conflict with the First Law
3. A robot must protect its own existence as long as such protection does not conflict with the First or Second Laws

Vanderelst and Winfield implemented these three laws in a humanoid programmable robot[2] (Vanderelst & Winfield 2018), and set up experiments where the robot's goals conflicted with the laws. In this experiment, the robot is controlled via a standard three-layer robot architecture: the top layer (controller) generates goals (e.g., "deliver package"), the second layer converts goals into tasks (e.g., "pick up object"), and the third converts tasks into sensori-motor executable actions (e.g., "move arm up"). This robot is supplemented with a fourth layer, that contains a consequence engine. The purpose of the consequence engine is to *predict* the state of the world, and the state of the robot, as a consequence of the action that the robot's controller plans to do. Any state of the world, or state of the robot, that led to a violation of any of the laws was encoded as a significant negative utility, with the first law violation having the most and the third law violation having the least negative utility. As the robot moved about in the world, trying to achieve its goals, it would invoke the consequence engine every few seconds. The consequence engine contained a model of the world, as well as a model of the robot itself, including the controller. It would evaluate the result of the robot's actions and evaluate whether the state of the world and/or the robot violated any of the three laws. Depending on any potential violations, the consequence engine would interrupt the robot controller's plan of action and create new goals that would attempt to minimize the negative utility experienced by the robot. The robot would try to achieve these new goals and return to the old goals once the new goals were met. However, if the world changed while the new goals were being achieved, the consequence engine could potentially create even newer goals that had then to be prioritized over the current goals, and so on. In summation, the consequence engine did not tell the robot *what to do*; rather it told the robot *what not to do*. The robot was then, experimentally, put in several situations where it had to continuously evaluate whether its goals, actions or even the actions of other entities in its world resulted in harm to humans or itself. The utility functions had to be carefully coded to ensure that the robot chose correctly, and in the absence of any dilemma, continued to achieve its functional goals. The programming of the utility functions makes the robot's *sense-of-ethics* extremely fragile. A small change in the code (even a syntactically correct typo) could completely reverse the robot's priorities.

[2] https://www.softbankrobotics.com/emea/en/nao

The difficulty in validating whether the robot's ethical sense was correct or not requires expensive and careful experimentation. However, there is no alternative to this expensive and careful experimentation to reassure us that the robot has a set of moral principles that guide its behaviour. For this reason, the field of implemented machine ethics currently leans towards using dilemmas as a validating whetstone (more on this in a later section) to check whether the implementation technique has succeeded or not.

## *Domains of Implementation*

All the attempts described previously have been concentrated in very few domains. This is a concern since the ethics that we expect out of autonomous vehicles are not the same as the ethical behaviour we expect out of autonomous healthcare robots. Although we are prepared to accept autonomous machines in specific domains, the notion of ethics is still considered to be a generalist concern. That is, an ethical machine is one that can interpret very general ideas from ethical philosophy and apply them to its specific domain. If we are to trust autonomous machines in multiple domains, then we must also concretize our ideas of what ethical behaviour in that domain means. This paper will discuss some of the implemented techniques, from the perspective of the domain that they have been positioned in.

### *Robotics and Cyber-Physical Systems*

In a presentation at the ACM annual conference in 2008, Arkin presented one of the first in-depth implementation attempts to embed ethical control and reasoning system in the field of autonomous weapons (Arkin 2008). The robots, in this case, are assumed to have a reactive/hybrid architecture where a deliberative mechanism was introduced to modulate the response that the robot makes. The intention behind such an effort was to enable a robot to obey the Laws of War and Rules of Engagement prescribed by international law. Robot control architecture typically uses mappings between stimuli and possible responses to decide how to act.

The architecture proposed by Arkin, provides options for an *ethical governor* (such that no unethical act is considered), an *ethical behavioural control* (unethical plans are constrained to generate ethical behaviour) or an *ethical adaptor* (transforms unethical actions onto ethical actions). While the architecture itself allows for flexibility in the reasoning engine used, and the ability of the robot to respond, it makes no recommendations on *how* the decision about the permissibility or impermissibility of an act is to be evaluated. While important in the implications of real-world impact by such systems, there are no experimental evaluations of robots deciding between multiple actions and how well (even post-facto) the robots fared in difficult situations.

Robots have been utilized in the healthcare industry, particularly in the care of the elderly (Moyle 2017). From assistive technologies such as exoskeletons, and robotic wheelchairs, to robots that elicit emotional responses (to act in a manner similar to animal-assisted therapy), there are a range of experiments using robots in this domain. A notable effort in this area is value-driven eldercare (Anderson et al. 2019) where the authors describe a healthcare robot called GenEth (short for *general ethical dilemma analyser*). GenEth works via encoding previously known and accepted principles in dealing with specific ethical dilemmas and then using Inductive Logic Programming to select a preferable action. GenEth uses a case-based representation that can generalize from previously presented cases and therefore deal with situations that are different from situations seen in training cases. GenEth is the latest in the authors' experiments in this domain, with MedEthEx (Anderson et al. 2006) and EthEl (Anderson & Anderson 2007) being the earlier iterations in a software-only form. This methodology of moving from software implementations to simulated robots, and finally real robots, presents a possible path for ironing out possible ethical 'bugs' before actual deployment in the real world.

Other studies in robot simulation (Lindner et al. 2017; Mackworth 2011) have used robots in simulated dilemmas, where one robot pretends to be a human while another robot's decision-making is tested. The simulated dilemma is usually a pre-decided dilemma, like the *Trolley problem* or *Cake/Death* dilemma (Armstrong 2015). The Trolley Problem is a dilemma created by Philippa Foot (Foot 1967), which seeks to clarify the problem of double-effect. That is, when an action has an intended good consequence, but also (unavoidably) has an unintended adverse consequence. In such cases, how should a machine/robot act? While these dilemmas are usually extreme cases, and unlikely to occur in real-life, they are viewed as benchmarks, which can be used by roboticists to demonstrate their system's reasoning capabilities (Bjorgen et al. 2018).

Functional imagination or a consequence simulation engine has been used as a way of testing the outcome of potential actions (Vanderelst & Winfield 2018). Inspired by the simulation theory of cognition (Marques & Holland 2009), the authors aimed to implement consequentialist ethics by having the robot imagine the future consequences of its actions, and then decide whether those consequences align with its goals. The authors implement an ethical layer that functions as a just-in-time checker of behavioural alternatives that are generated by the robot's controller. The ethical layer uses a simulation module to predict the future sensory and motor states of humans as well as itself. These are then evaluated using an evaluation module to test if any particular state might be undesirable. For instance, if a robot acts only to achieve its goal, it might put the human in physical danger (say, by pushing it). Now, if the robot could simulate the future caused by its action, the evaluation module would point out that a human would be harmed by its action, which would be then forbidden by the ethical layer. The authors used Asimov's Three Laws of Robotics to test their robots' ability to avoid (or delay) orders that could potentially harm human beings or themselves.

In an evaluation of health-care robots assisting carers who work with patients that have Parkinson's Disease, some robots were augmented with an Intervening Ethical Governor (Shim & Arkin 2017). This was a follow-on from previous work on Ethical Governors in robots (Arkin 2008), and used deontological ethics to achieve ethical behaviour. Rules regarding obligatory and prohibited behaviours were encoded into intervention procedures. The robot evaluated whether certain perceptual states, by the patient or the carer, violated any prohibited behaviour rules and triggered an intervening action. It would also autonomously generate interventions if the patient or caregiver violated any obligatory rules. For instance, if the robot detected prohibited behaviours (*e.g.,* the patient was yelling or using foul language) it would generate an intervention action based on medical guidelines and expert reviews. The robot was able to prioritize the safety of the patient over other obligation rules and generate actions in the face of multiple stimuli.

Unmanned Autonomous Vehicles (UAVs) are a good example of systems that already have a considerable amount of autonomy, and will continue to increase their autonomic capabilities in the near future. Model-checking and verification of a UAV's planned actions have been tested against common-sense dilemmas (Dennis et al. 2016) to ensure that the agent chooses the least unethical action in case a dilemma arises. However, as the amount of autonomy increases, the range of activities UAVs undertake also increases. This implies that the context in which the UAV operates, and the policies allowed, will change dynamically. This dynamic change cannot, currently, be handled by model-checkers in real-time. It was reported that verifying four properties for their UAV took four days to complete (Dennis et al. 2016).

### *Normative Agents and Value Frameworks*

There have been some works that approach the problem of implementing machine ethics from a different perspective. Instead of trying to make existing software systems/robots have a mechanism for sensing and reasoning about the ethics of a situation or behaviour, these approaches attempt to create a computational framework to represent ethics themselves. That is, the focus is not on implementing a specific theory, but rather to build a computational framework for representing any moral value.

In many situations, there will be more than one agent that acts and more than one perspective on which acts are ethical. In such situations, agents must have the ability to represent and evaluate, not only their own behaviour, but also other agents' behaviour. This is achieved via an explicit representation of theories of good and theories of right along with an agent's ethical preferences. Combined with a judgement process, the agent can generate possible combinations of actions that satisfy all the constraints of the moral values it has been given. The authors use Answer Set Programming (a form of declarative programming which can represent knowledge-intensive problems and search through possible solutions very quickly) to create BDI (*belief-desire-intention*) agents that can reason about the priorities between desirable and moral actions (Cointe et al. 2016).

Computational models of ethical theories often embed ethical decision-making directly within an agent's decision-making process, thus making it very difficult for the agent to infer cases and reason about its behaviour. The authors present an Event calculus (a logical formalism that allows representation of events and their effects) that allows agents to create causal traces of actions and their consequences (Berreby et al. 2018). This work builds upon work to create a higher-level action language to create autonomous agents that can reason about ethical behaviour (Cointe et al. 2016).

In a different approach, instead of explicitly implementing ethics, agents were programmed to select norms autonomously, as an optimisation problem (Serramia et al. 2018). The authors view the choosing of norms as an optimization problem, given a set of constraints and preferences. They express norms as a pair, which connects an agent with a set of actions, with deontic operators of permission, obligation, or prohibition. They utilize three norm relationships of exclusivity, substitutability, and generalisation to generate a norm system that is both conflict-free as well as non-redundant. The approach then uses multi-objective optimisation to satisfy all the constraints and achieve as many preferences as possible. Moral values are considered as a set of values, with each norm supporting some subset. The problem of upholding some moral values is now reduced to selecting the smallest set of norms, that supports all the moral values we care about. If these norms are encoded as linear programs, then the best set of norms can be calculated by a linear program solver.

One of the problems that an autonomous agent could have, is that it might resist any change of behaviour, or might encourage changes to its own rules that might not be ethical. Agents that 'learn on the job' are value-loading or value-selecting agents, and have the potential to prevent ethical governors or utility-based rules from enforcing ethical behaviour (Armstrong 2015). The authors introduce the *Cake or Death* dilemma (explained below) to illustrate the nature of such an agent and propose a meta-utility function that mediates how the value-selecting agent can change its utility function without introducing artificial resistance or encouragement in the process.

It is difficult to be completely definitive about which implementation technique is being used by an approach (esp. when not all implementations are available for open-source scrutiny). In **Error! Reference source not found.**, we have inferred the mechanisms used by some implementations; however it is not always clear whether the action representation is independent of the ethical representation, or whether the two are inextricably linked. The entry called **hybrid**, in **Error! Reference source not found.**, refers to the fact that the authors have used multiple techniques (Inductive Logic Programming, Case-based reasoning) for deriving ethical rules and then perform reasoning using these rules.

**Table 1: Implementation Techniques attempted for ensuring ethical behaviour**

| Domains / Implementation Techniques | Robots and Robot Simulations | UAVs or Cyber-Physical Systems | Software-Only Systems |
|---|---|---|---|
| **Rule-Based** | Bringsjord et al. 2006; (Vanderelst & Winfield 2018) | Dennis et al. 2016 | |
| **Constraint-Satisfaction** | Arkin 2008; Mackworth 2011 | | Cointe et al. 2016 |
| **Reinforcement Learning** | | | Abel et al. 2016 |
| **Causal Networks** | Lindner et al. 2017 | | |
| **Normative Agents** | | | Serramia et al. 2018; Cointe et al. 2016; Armstrong 2015 |
| **Hybrid** | (Anderson et al. 2019) | | |

## *Evaluation Using Dilemmas – A Challenge to Ethicists*

The most popular mechanism of evaluation is by simulation of ethical dilemmas. From an implementation and engineering perspective, such mechanisms are critical to creating trust in ethical machines. Regardless of the method of implementation (constraint satisfaction/machine learning/rule-based methods etc.) or the ethical theory being implemented, ethical dilemmas offer a 'neutral' and 'objective' way of testing the performance of an ethical machine. The common set of dilemmas being used is given below:

1. *The Trolley Problem*: First proposed by Philippa Foot (Foot 1967), the problem is a thought-experiment that seeks to discuss the issues raised by actions that have double effects. That is, some actions may have good intentions/goal behind them, however the means used to achieve the goal may themselves be regrettable or even reprehensible. A common formulation of this dilemma is given as: There is a runaway trolley which can only be steered between two tracks.

Five men are working on one track, and one man is working on the other one, and depending on the track that the trolley is on, people will die. The question is which track should the driver of the trolley steer the track on to? This is the most popular dilemma used in machine ethics implementation evaluations.

2. *The Burning Room*: This dilemma was proposed by (Abel et al. 2016), in turn based on a dilemma proposed by (Briggs & Scheutz 2015), that illustrates whether an artificial agent should attempt to sacrifice itself to preserve something of value to the human. If the human's valuation of an arbitrary object, relative to the artificial agent, was completely known then the agent could take definitive action. However, in the absence of information and lack of time to obtain said information, the artificial agent is unable to decide on a course of action.
3. *Cake or Death*: This dilemma deals with the supposedly easy choice of whether a robot should choose to bake a cake or kill someone. Proposed by Stuart Armstrong (Armstrong 2015), it shows how, depending on the formulation of the problem, and the selection of values, an agent could consistently reach the obviously wrong solution. That is, an agent can manipulate its own value framework to rationally avoid the hard choices that an ethical agent is actually required to make.
4. *The Lying Dilemma*: This dilemma is another instance that illustrates the doctrine of double effect. As discussed in Lindner et al. 2017, this dilemma concerns a health-care robot that discovers lying and the use of guilt to encourage its patient to adopt healthier living habits, such as exercise. The question is when would it be acceptable to lie in order to achieve a good outcome?

This set of dilemmas may be criticized as only illustrating a particular kind of problem in decision-making or being applicable only in limited domains. This set may also be illustrative of the biases of technologists who implement them. A simple solution to the limited domain objection would be to increase the set of dilemmas, to accommodate a broader set of domains where ethical machines will operate. The larger objection to dilemmas being an unsuitable technique of evaluation is more difficult to reconcile with engineering. Engineers and computer scientists feel the fundamental need for benchmarks or other objective evaluation mechanisms. In the absence of any alternative proposals for objective evaluation, the use of dilemmas as a validating mechanism for implementing machine ethics will continue.

## Characterizing Machine Implementations

There is no consensus among technologists about the best method to implement ethics, best mechanism to validate implementations, or even the best ethics *to implement* in machines. This is not surprising since there are no generally accepted test cases that allow implementations to be benchmarked. Hence we propose the following axes to compare various implementations across domains and ethical theories:

1. *Evaluation Using User Studies (EU)*: The ethical implications of autonomous systems' behaviour come to the fore when they are evaluated in situ, or in actual human environments
2. *Evaluation Using Simulations (ES)*: User-studies are expensive to conduct on a large scale. Simulations can be used to perform a first-level evaluation of the implementation
3. *Action Representation (AR)*: How does the autonomous system represent its own and others' actions, and reason about them?
4. *Ethics Representation (ER)*: How does the autonomous system represent the ethical values/rules that it seeks to uphold?
5. *Single Agent / Multiple Agents (SMA)*: Does the autonomous system assess a single agent's point of view, or does it assume that multiple agents with differing capabilities/goals exist in the interaction?
6. *Continuous Learning (CL)*: Any autonomous system that is long-lived must adapt itself to the humans it interacts with. All social mores are subject to change, and what is considered ethical behaviour may itself change. Although this is not an immediate problem (most systems are not long-lived enough), it could be significant in deploying the same system among different communities.

7. *Open / Downloadable Implementation (OI)*: Trusting autonomous systems to behave in an ethically acceptable manner, requires independent validation of the claimed behavioural stance of the machines.

We present a comparison of the implementations discussed in the previous section, in Table 2.

**Table 2: Comparing Ethics Implementations**

| | EU | ES | AR | ER | SMA | CL | OI |
|---|---|---|---|---|---|---|---|
| (Anderson et al. 2019) | Yes | Yes | Sensor data represented as a boolean values in a PerceptionList. Values are compared with a DutySet to find duty satisfaction/violation values. | Human-ethicist provided cases are learnt via a decision-tree of PerceptionList and outcome pairs. Depending on duty satisfaction/violation values, actions are sorted in order of ethical preference. | Multiple single agents, in turn | No | Yes |
| (Vanderelst & Winfield 2018) | No | Yes | Three-layered model for robot control, along with a simulation engine for predicting future sensory and internal states of robot and human. | Evaluation module for converting simulated human and robot states into desirability metrics. Consequential ethics modelled by Asimov's Laws of Robotics | Single | No | Yes |
| (Berreby et al. 2018) | No | Yes | An extended form of event calculus, containing agents, timepoints, actions, omissions, fluents, events, simulations etc. along with an EventMotor and a PlanningContext. | Causality modelling using tree of simulations to establish properties of counter-factual validity, cruciality and necessity | Multiple | No | Yes |
| (Serramia et al. 2018) | No | No | Norms encoded as a triad of a deontic operator (permission, obligation, prohibition), an agent and an action. Norms have relationships between them of exclusivity, substitutability and generalisability | The set of Norms that are chosen by some utility function represents the moral values that are encoded for that agent society. Encoding of Norms as a linear program, along with preference criteria and constraints allows for optimal norm system to be chosen | Multiple | No | No |
| (Shim & Arkin 2017) | Yes | No | Perceptual data mapped onto logical assertions which form an EvidenceSet | No particular theory is implemented. Two prohibition and obligation rules from medical literature were created. If EvidenceSet triggered a rule, an intervention occurred. | Multiple single agents, in turn | No | No |
| (Lindner et al. 2017) | Yes | Yes | Actions and consequences are represented as acyclic graphs, within a causal agency model that uses Pearl-Halpern-style causal networks | Ethical principles are modelled as different evaluations of causal agency models. The consequences of the models are evaluated by applying the ethical principle encoded as a logical formula. A model checker returns | Single agent | No | Yes |
| | | | | whether a particular principle would allow a particular action (mapped to a model) is permissible or not. | | | |
| (Dennis et al. 2016) | No | No | Plans are sets of actions that lead towards goals. While executing a plan, the beliefs and goals of an agent may change, as the environment changes. | An ethical principle is a finite set of propositional logic formulae. An ethical policy defines a total order on the formulae. Model-checking ensures that during plan selection, preferences among the formulae are observed. If plans conflict in principles that are violated, then the least unethical plan is chosen | Multiple agents | No | No |
| (Cointe et al. 2016) | No | No | In the context of beliefs, desires and intentions, every agent has an evaluation process that describes an action as a tuple--pair of conditions and consequences. Since consequences affect beliefs and desires, the evaluation process produces executable and desirable actions. | Given a set of desirable, and feasible actions, an ethical principle is a function that represents a philosophical theory and evaluates if any constraints are violated by the given set of actions. Implemented using AnswerSetProgrammin g | Multiple agents | No | Yes |
| (Abel et al. 2016) | No | Yes | The world, possible actions, change of environment, probabilities of change, are collectively modelled as a Partially Observable Markov Decision Process (POMDP). The POMDP is then 'solved' to achieve ethical actions in all possible transitions. | An ethical principle is encoded as a hidden utility function in the environment, which must be learnt by the RL agent. | Single agent | No | Yes |
| (Armstrong 2015) | No | No | Agents are utility maximizing beings, that given some evidence and set of actions, will choose to perform that action which will lead to a world which yields the highest utility | Compound utility functions relating possible worlds, and their associated utilities are given to the agent. The agent tries to update its knowledge of the evidence, and then choose actions in such a manner that the value function over the achieved world is maximized. | Single agent | Ye s | No |

**See Appendex I for Open / Downloadable Implementation URLs for various approaches.**

### *Which Implementation To Use*

Implementing a particular ethical theory in an autonomic system reassures users and society, that the machines that affect their lives have some set of morals, that have previously been examined by experts. The alternative is to depend on the system designer or programmer to ensure that their creations will always be benign, which may be impossible in many situations. Some implementations have demonstrated that their mechanism can handle multiple types of ethical theories, while others have stuck to a single one, with arguments based on their domain. Regardless of whether an implementation is flexible with ethical theory or not, a more critical question is which particular ethical theory *should* be used for machine implementations. Among moral philosophers themselves, there is little agreement on which particular theory could and should be implemented both by human and machine agents (Bogosian 2017). This moral uncertainty is not necessarily a bad thing since different theories might plausibly be the correct theory to follow, in different situations. According to MacAskill, in real life, normative uncertainty is the norm, and hence we should aim to maximize *expected choice-worthiness* under normative uncertainty in machines, as well (MacAskill 2016). MacAskill puts forward a voting framework that attempts to perform an inter-theoretic comparison of different ethical theories for different situations, taking into account the user's credence in each theory.

## Conclusion

This paper reports on the many attempts at ethics by design in multiple domains, as well as by using multiple techniques. However, none of these have yielded a result that is satisfactorily robust in multiple situations or is immune to designer-bias. In this author's opinion, regardless of the implementation technique, the same machine must be able to handle different contexts, by simply updating credence in theories, while being consistent intra-context. The need for different contexts is easily seen by the number of domains that autonomous machines are being introduced into. As the number of domains increases, the need for domain-specific and robust ethical standards increases in urgency. Specifically, the notion of explainability, in the author's opinion, must be built into the mechanism, since no machine however perfect will be trusted, if it cannot explain its decisions. The resulting ethics should also be flexible in dealing with multiple different situations and must survive competition with other machines that may not have the same set of ethical standards.

## Appendix I: Open / Downloadable Implementation URL

- (Anderson et al. 2019) https://www.researchgate.net/publication/333999191_GenEth_Distributionzip
- (Vanderelst & Winfield 2018) - Not Found At Time of Writing
- (Berreby et al. 2018) https://github.com/FBerreby/Aamas2018
- (Lindner et al. 2017) http://www.hera-project.com/software/
- (Cointe et al. 2016) - Not Found At Time of Writing
- (Abel et al. 2016) https://github.com/david-abel/ethical_dilemmas